\documentclass[twoside,letterpaper,journal]{IEEEtran}
\usepackage{amsmath,amsfonts}
\usepackage{algorithmic}
\usepackage{algorithm}
\usepackage{array}
\usepackage[caption=false,font=normalsize,labelfont=sf,textfont=sf]{subfig}
\usepackage{textcomp}
\usepackage{stfloats}
\usepackage{url}
\usepackage{verbatim}
\usepackage{graphicx}
\usepackage{cite}
\usepackage{pifont}
\usepackage{colortbl}
\usepackage{multirow}
\usepackage{orcidlink}
\hypersetup{hidelinks}
\definecolor{lightgray}{gray}{0.85}

\begin{document}

\title{Decoupling Spatio-Temporal Adapter for\\ Fine-Grained Badminton Action Localization}

\author{Tianyu Wang\textsuperscript{\orcidlink{0000-0001-6250-3656}}, \and Junjie Wu\textsuperscript{\orcidlink{0000-0001-7650-3657}}, \and Jingquan Gao\textsuperscript{\orcidlink{0009-0001-7662-8158}}, \and Shishuo Li\textsuperscript{\orcidlink{0009-0002-6999-4739}}
\thanks{Corresponding author: Shishuo Li (e-mail: sy2408214@buaa.edu.cn).}
\thanks{Tianyu Wang, Jingquan Gao, and Shishuo Li are with the School of Economics and Management, Beihang University, Beijing 100191, China.}%
\thanks{Junjie Wu is with the School of Economics and Management, Beihang University, Beijing 100191, China, and also with the Key Laboratory of Data Intelligence and Management, Beihang University, Ministry of Industry and Information Technology, Beijing 100191, China.}
\thanks{This work was supported by the National Natural Science Foundation of China under Grant 72201015.}
}

\markboth{IEEE Transactions on Multimedia}%
{Wang \MakeLowercase{\textit{et al.}}: Decoupling Spatio-Temporal Adapter for Fine-grained Badminton Action Localization}

\IEEEpubid{}

\maketitle

\begin{abstract}
Temporal Action Localization (TAL) has been extensively studied in generic video understanding, while fine-grained sports scenarios, such as professional badminton, remain underexplored due to their complex and subtle spatio-temporal dynamics. In this paper, we focus on fine-grained TAL in professional badminton videos and introduce a new benchmark dataset, Fine-Badminton, which consists of 31 matches with 29 fine-grained stroke categories, covering 2104 rallies and 27597 annotated actions. To effectively capture the intricate motion patterns in such scenarios, we propose a Decoupling Spatio-Temporal Adapter (DSTA), which enables efficient modeling of spatio-temporal features within a parameter-efficient framework. Specifically, DSTA decomposes motion representation into three parallel branches, capturing temporal dynamics as well as vertical and horizontal spatial variations. The design allows the model to better distinguish subtle differences among fine-grained actions. Extensive experiments on both the Fine-Badminton dataset and the ShuttleSet benchmark demonstrate that the proposed method achieves state-of-the-art performance while introducing only a marginal increase in computational and parameter cost. These results validate the effectiveness and efficiency of the proposed approach for fine-grained temporal action localization.
\end{abstract}

\begin{IEEEkeywords}
Sport video, temporal action localization, temporal modeling.
\end{IEEEkeywords}

\section{Introduction}
\label{sec:introduction}
\IEEEPARstart{T}{he} exponential growth of video data has exceeded the processing capacity of manual analysis systems, driving research interest in video understanding \cite{2020mmaction2, 2024opentad, cheng2022tallformer, ETAD, adaTad, mmcv, slowFast, BSN, BMN, detr1, detr2, detr3, videomae, actionFormer, MREM, ada_improve_3}. Among these tasks, Temporal Action Localization (TAL) \cite{cheng2022tallformer, 2020mmaction2, BSN, BMN, ETAD, adaTad, 2024opentad, detr1, detr2, detr3, ada_improve_3, MREM}, which aims to locate actions within the video, is particularly challenging as it requires fine-grained, frame-wise understanding. 

In this paper, we focus on TAL in professional badminton scenarios. As summarized in Table~\ref{tab:1}, there exist badminton datasets that include ShuttleSet \cite{ShuttleSet} and Badminton Olympic \cite{badminton-dataset-towards-olympic}. ShuttleSet lacks temporal interval annotations, limiting frame-wise evaluation; Badminton Olympic adopts a coarse-grained label set, making it unsuitable for fine-grained TAL. To address this gap, we introduce Fine-Badminton, a fine-grained professional badminton video dataset with frame-wise annotations. The videos are collected from platforms such as YouTube, including 31 videos in 29 fine-grained categories (\emph{e.g.}, "Lift" and "Drop Shot"). Expert annotators provide precise frame-wise timestamps for each action.

\begin{figure}[!t]
\centering
\includegraphics[width=\linewidth]{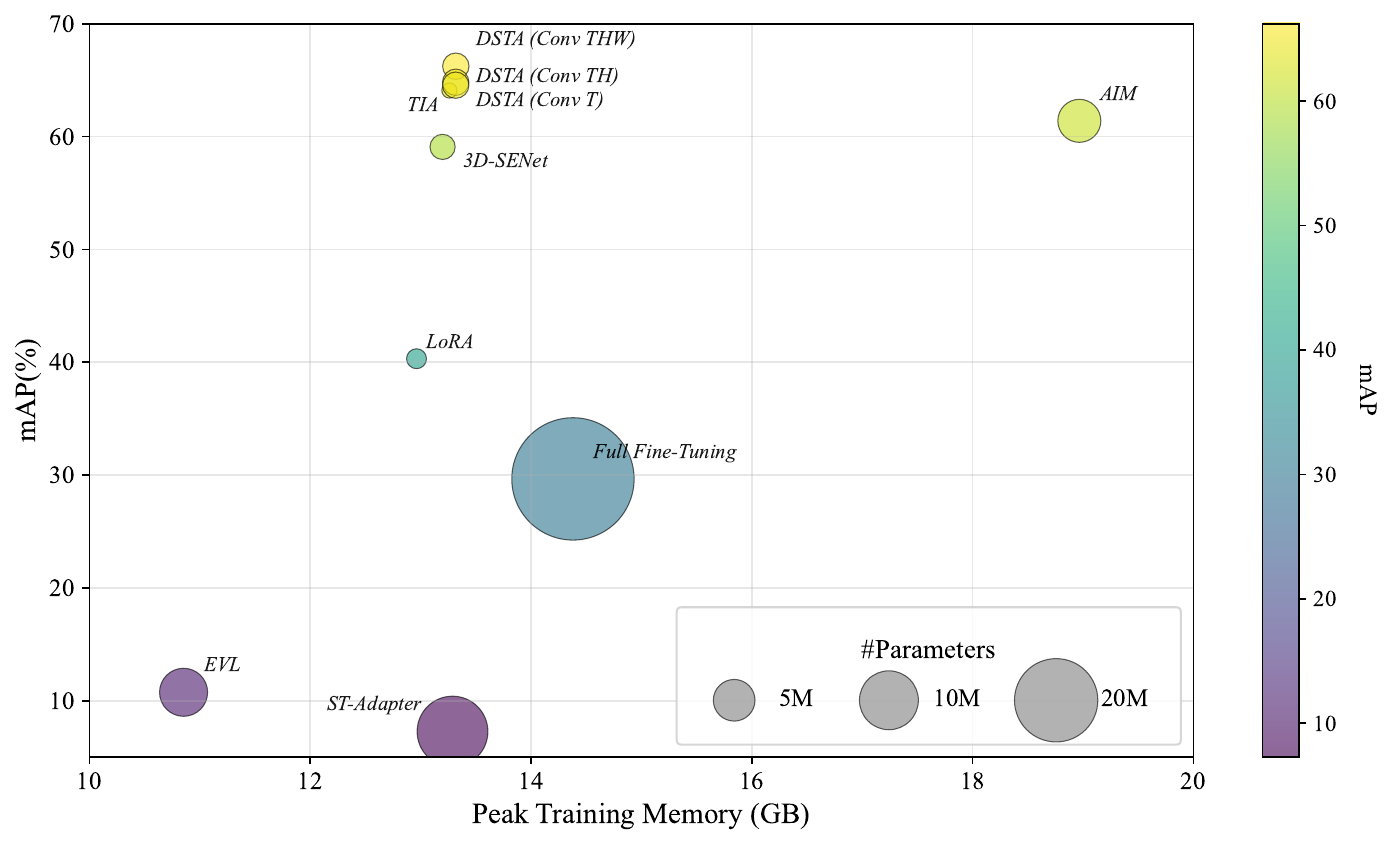}
\caption{The memory and performance of TAL models with different transfer learning methods.}
\label{fig:memory and performance}
\end{figure}

Compared to generic TAL datasets \cite{ActivityNet, THUMOS14}, our Fine-Badminton dataset presents two challenges: 1) It involves environments with diverse camera viewpoints and badminton venues, increasing the difficulty of temporal localization; 2) Badminton actions are characterized by short duration, high speed, and subtle motion, requiring stronger modeling of local spatio-temporal features. In this context, our goal is to develop a cost-effective TAL model with accurate spatio-temporal modeling.

\begin{table*}[!t]
\begin{center}
\caption{The statistical characters of several representative TAL datasets.}
\label{tab:1}
\begin{tabular}{lcccccccccc}
\hline
Dataset & Domain  & Interval Annotation & Fine-grained & \#Annotations  & \#Classes & Year\\
\hline
ActivityNet-1.3 \cite{ActivityNet}  & General & $\checkmark$ & \ding{55} & 30800 & 200 & 2015 \\
    THUMOS14 \cite{THUMOS14} & General & $\checkmark$ & \ding{55} & 6365 & 20 & 2014 \\ 
    Badminton Olympic \cite{badminton-dataset-towards-olympic} & Badminton & $\checkmark$ & \ding{55} &  15327 & 12 & 2017 \\ 
    BadmintonDB \cite{BadmintonDB}& Badminton & $\checkmark$ & $\checkmark$ & 9671 & 10 & 2022\\
    ShuttleSet \cite{ShuttleSet} & Badminton & \ding{55} & $\checkmark$ & 36492 & 18 & 2023\\
    VideoBadminton \cite{VideoBadminton} & Badminton & $\checkmark$ & $\checkmark$ & 7822 & 18 & 2024\\
    \textbf{Fine-Badminton} (Ours) & Badminton & $\checkmark$ & $\checkmark$ & 27597 & 29 & 2026\\
\hline
\end{tabular}
\end{center}
\end{table*}

\IEEEpubidadjcol

A recent Temporal-Informative Adapter (TIA) \cite{adaTad} integrates temporal depth-wise convolutions (DWConv) to aggregate local contextual information from adjacent frames with low memory cost. However, it mainly focuses on temporal aggregation while neglecting directional spatial dynamics along vertical and horizontal dimensions, which are important for fine-grained badminton action discrimination. To address this limitation, we propose the Decoupling Spatio-Temporal Adapter (DSTA), a three-branch architecture with separate convolutional kernels for temporal, vertical, and horizontal modeling. The vertical and horizontal branches capture directional motion patterns, such as smashes and net shots, while retaining the original DWConv for temporal modeling. As shown in Fig.~\ref{fig:memory and performance}, DSTA achieves better performance than TIA \cite{adaTad} with only marginal additional memory cost.

To validate the effectiveness of our proposed method, we evaluate it on ShuttleSet \cite{ShuttleSet} and Fine-Badminton. Following the THUMOS14 \cite{THUMOS14} evaluation protocol, our model achieves a mean Average Precision (mAP) of 74.67\% on ShuttleSet and 66.23\% on Fine-Badminton, outperforming existing adapter-based methods. 

We summarize our main contributions as follows: 
\begin{enumerate}
    \item We introduce a fine-grained badminton dataset for temporal action localization, releasing a representative subset of 10 annotated videos with frame-level temporal annotations across 29 action categories on Zenodo \cite{fine-badminton2026} as a benchmark resource for this task.
    \item We propose a decoupling spatio-temporal modeling strategy and design a novel adapter, termed DSTA, to enhance spatial motion representation.
    \item Extensive experimental results demonstrate that DSTA consistently outperforms existing adapter-based methods on both the public ShuttleSet dataset \cite{ShuttleSet} and our Fine-Badminton dataset.
\end{enumerate}

\section{Related Works}
\label{sec:Related Works}
\subsection{Temporal Action Localization}

Temporal Action Localization (TAL) aims to identify specific actions within video sequences and accurately determine their temporal boundaries.
The emergence of deep learning techniques based on attention mechanisms has enabled effective modeling of action-related spatio-temporal information, thus improving the accuracy and robustness of TAL.
Depending on the architectural design and the stages at which temporal boundaries are determined, and action classification is performed, TAL models can be divided into three categories: single-stage methods, two-stage methods, and DETR-based methods. 
Single-stage methods such as BMN \cite{BMN} and ActionFormer \cite{actionFormer} simultaneously perform action classification and temporal boundary regression by using multi-scale feature pyramids to directly predict action instances.
Two-stage methods such as MREM \cite{MREM} and BSN \cite{BSN} first regress temporal boundaries and subsequently perform action classification.
A key distinction between DETR-based methods \cite{detr1,detr2,detr3} and other approaches is their use of learnable queries to directly predict the start and end times of target actions, instead of relying on proposal-based mechanisms. 
Some TAL models can be equipped with backbone networks for video feature extraction, such as VideoMAE \cite{videomae} and SlowFast \cite{slowFast}.
Based on the overall processing pipeline, the models can be divided into two categories: feature-based methods and end-to-end models.
Feature-based methods rely on pre-extracted visual features, sometimes combined with optical flow features. For instance, BSN \cite{BSN} utilizes a pretrained two-stream network to extract spatial appearance features and temporal motion features.
In contrast, end-to-end methods integrate feature extraction and temporal action localization into a unified learning framework. 
The Tallformer model represents an advancement in this direction by directly operating on RGB inputs and introducing long-range temporal memory mechanisms.

To improve spatio-temporal representation learning in end-to-end video models, several studies have explored decomposed convolutional architectures. R(2+1)D \cite{tran2018closer} factorizes 3D convolution into spatial and temporal components, while S3D \cite{xie2017rethinking} adopts separable 3D convolutions for efficiency. Inspired by these approaches, our DSTA extends this idea to parameter-efficient adapters by introducing temporal, vertical, and horizontal branches for directional motion modeling.

\begin{figure*}[!t]
    \centering
    \subfloat[Standard]{\includegraphics[width=0.13\linewidth, trim=0 0 0 10, clip]{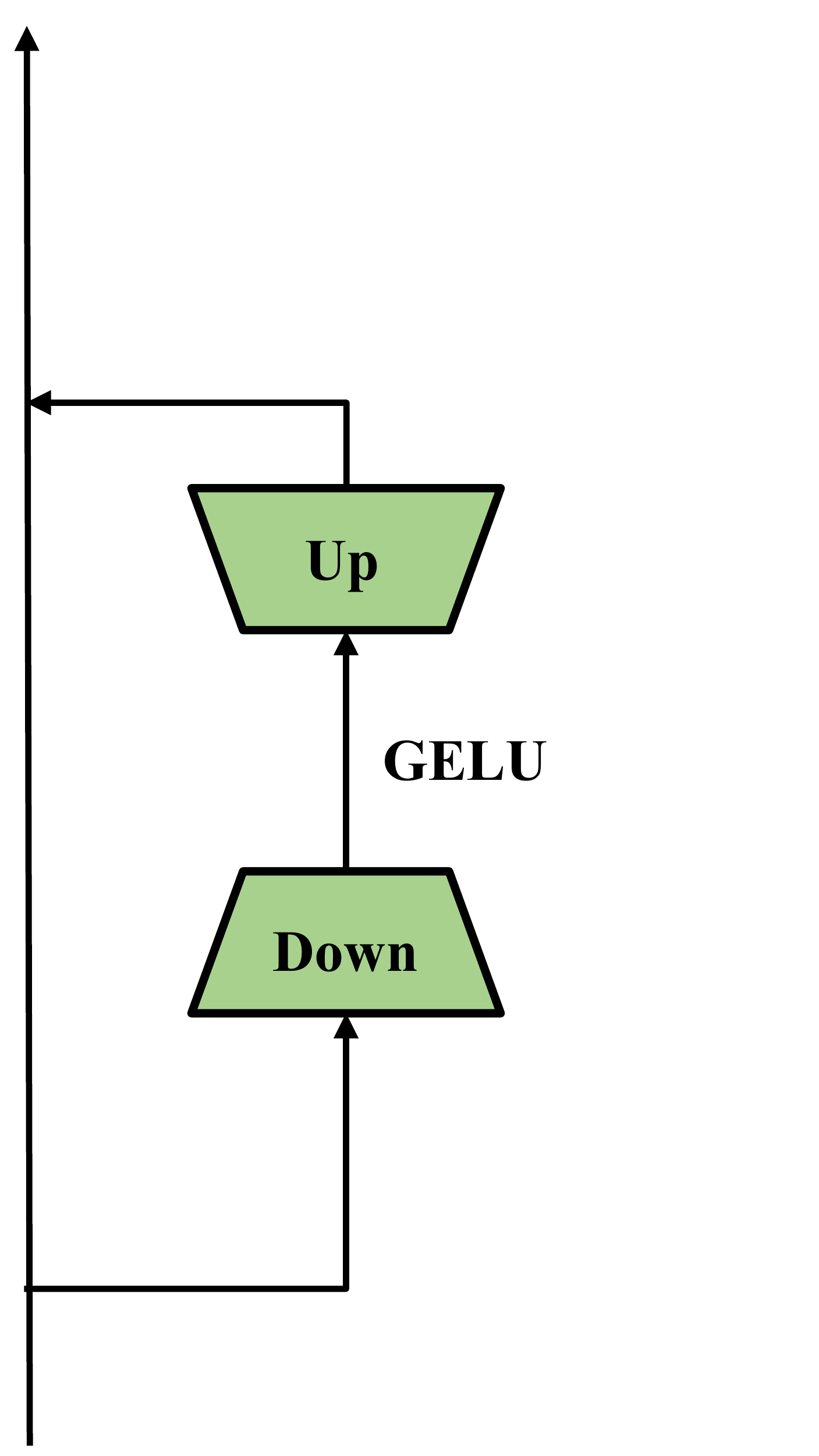}\label{fig:standard adapter}}%
    \hfil
    \subfloat[TIA]{\includegraphics[width=0.13\linewidth, trim=0 0 0 10, clip]{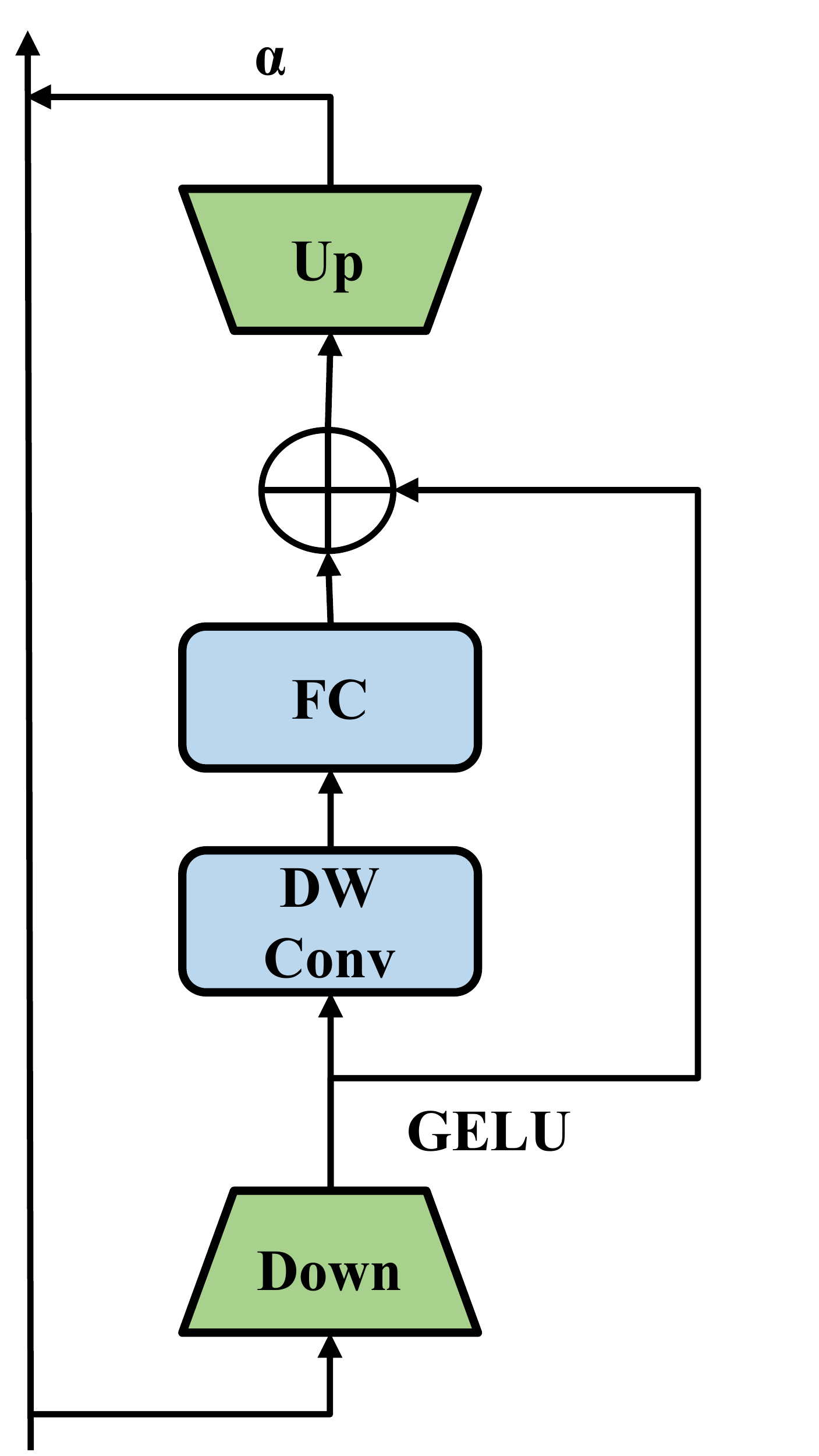}\label{fig:TIA}}%
    \hfil
    \subfloat[3D-SENet]{\includegraphics[width=0.13\linewidth, trim=0 0 0 10, clip]{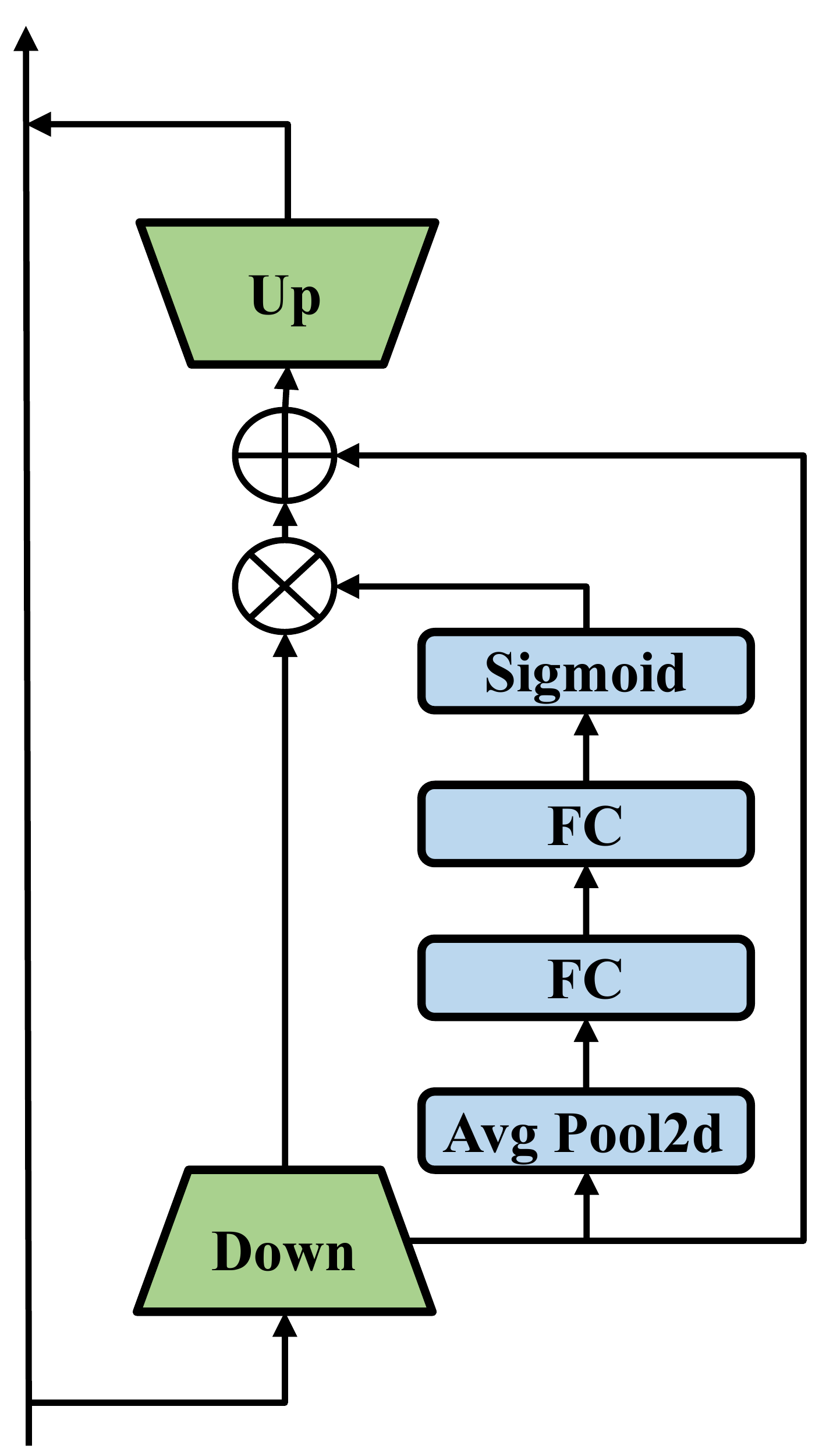}\label{fig:adapter_3dsnet}}%
    \hfil
    \subfloat[DSTA]{\includegraphics[width=0.13\linewidth, trim=0 0 0 10, clip]{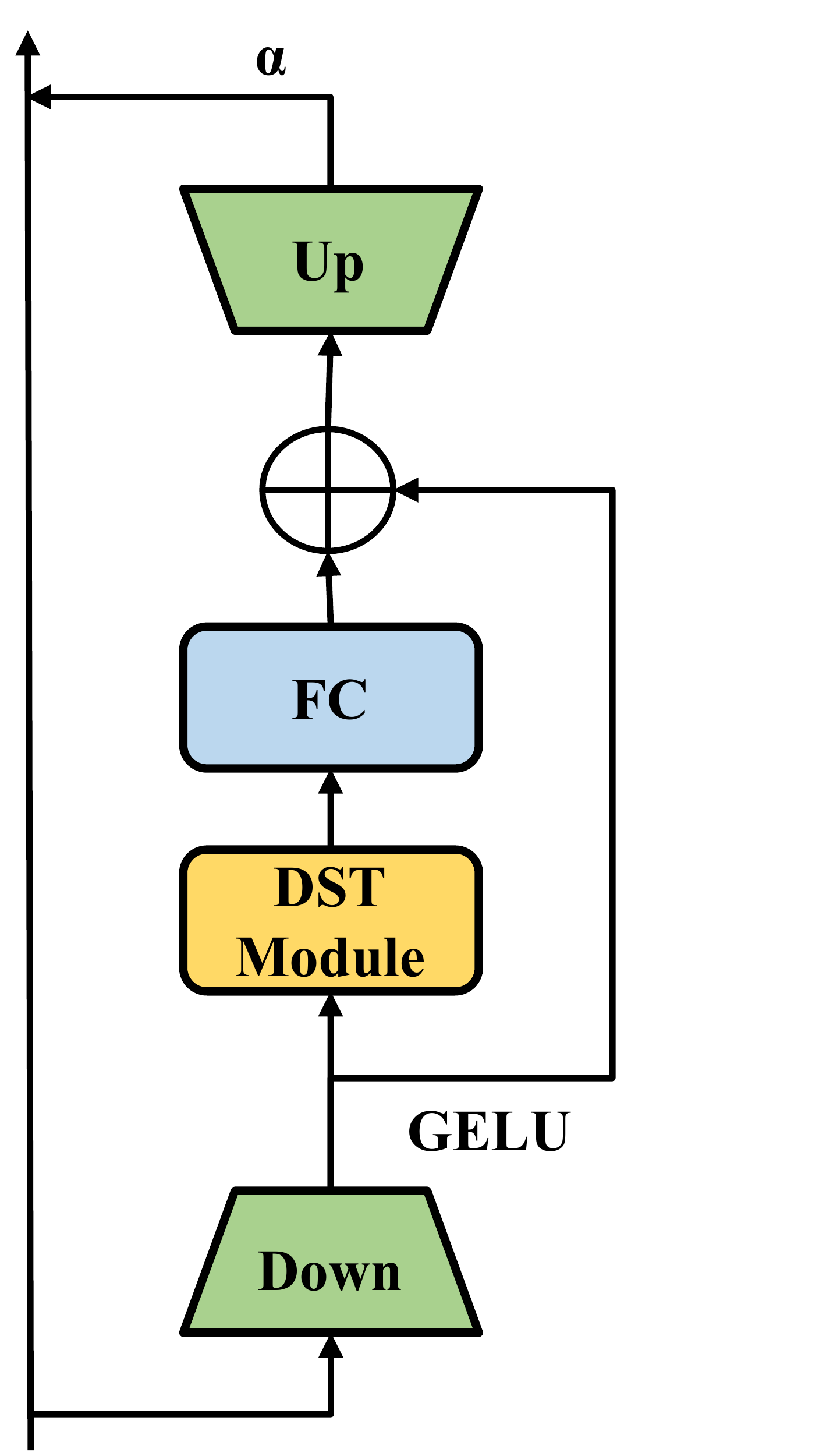}\label{fig:DSTA}}%
  \caption{Comparison of different adapter architectures:
    (a) Standard Adapter,
    (b) TIA,
    (c) 3D-SENet Adapter,
    and (d) DSTA (ours).} 
  \label{fig:adapter}
\end{figure*}

Research in natural language processing \cite{originadapter} proposes a transfer learning approach using adapter modules, as illustrated in Fig.~\ref{fig:standard adapter}. 
Adapter modules are lightweight and scalable components that introduce only a small number of trainable parameters for each downstream task, and new tasks can be added without modifying previously learned parameters. 
This approach preserves the original network parameters and achieves parameter sharing. Considering the trends of end-to-end training and model scaling, AdaTAD \cite{adaTad} significantly reduces the adaptation cost of backbone models by inserting a Temporal-Informative Adapter(TIA) based on the standard adapter architecture \cite{originadapter}, as shown in Fig.~\ref{fig:TIA}. 
It focuses on temporal features and exhibits limited capability in modeling spatial features.
TIA has been integrated into the encoder of VideoMAE \cite{ada_improve_1}, as well as into models in the WEAR repository \cite{ada_improve_2}.
As shown in Fig.~\ref{fig:adapter_3dsnet}, 3D-SENet \cite{ada_improve_3} further enhances spatio-temporal feature aggregation through attention mechanisms by modifying the structure of TIA.

\subsection{Badminton Video Dataset}
\label{subsec:badminton_video_dataset}

There are a few publicly available badminton datasets, and some of the more popular ones include BadmintonDB \cite{BadmintonDB}, VideoBadminton \cite{VideoBadminton}, Badminton Olympic \cite{badminton-dataset-towards-olympic}, and ShuttleSet \cite{ShuttleSet}.
BadmintonDB \cite{BadmintonDB} is a dataset comprising nine real-world badminton matches between two players, consisting of 811 rallies and 9671 strokes. Each record provides the start and end times of rallies, strokes, and point outcomes, along with 10 standard shot types defined by the BWF coaching manual and distinguished by forehand and backhand variations. The dataset also includes five categories of rally outcomes: fault, out, miss, misjudge, and lucky. This dataset is designed for player-specific tactical analysis and real-time rally outcome prediction; however, it categorizes badminton strokes into only 10 coarse types, differentiated by forehand and backhand.
VideoBadminton \cite{VideoBadminton} focuses on action recognition, where each video corresponds to a single annotated action. 
The action labels for this dataset include block, cross-court flight, cut, defensive clear, defensive drive, drop shot, flat shot, lift, long serve, push shot, rear-court flat drive, rush shot, short flat shot, short serve, smash, tap smash, and transitional slice.
In the Badminton Olympic \cite{badminton-dataset-towards-olympic}, the time boundaries and labels of strokes, point-winning moments, and the positions of players in selected frames are annotated.
ShuttleSet\cite{ShuttleSet} has annotations of 3685 rallies and 36492 strokes, which can be used for TAL.
The action labels for this dataset include back-court drive, clear, cross-court net shot, defensive return drive, defensive return lob, drive, driven flight, drop, lob, long service, net shot, passive drop, push, return net, rush, short service, smash, and wrist smash.
This is consistent with Badminton World Federation (BWF) standards but is relatively less intuitive to the broader badminton community.

\subsection{Badminton Video Analysis}
The main research on classical vision tasks in badminton includes object detection and tracking, action recognition, and temporal action localization.
Models such as TrackNet \cite{tracknetv2,trackball_3_tracknetV3} take the design of the U-Net and focus on the detection of the ball.TrackNetV3 further introduces a rectification module, which achieves precise positioning through post-processing.
MonoTrack \cite{Liu2022MonoTrackST} is designed to generate and integrate visual sub-signals, including court layout, stance, and badminton position, in order to segment unlabeled video into a known point-and-shoot loop structure and to produce 3D trajectories of the shuttlecock that can be used for tactical analysis.
The work on detecting players is often derived from analyzing the controllable areas of players \cite{trackplayer_1}.
ShuttleNet \cite{wang2021shuttlenetpositionawarefusionrally} is a model built specifically for badminton, which relies on large-scale annotated training data, to enable TAL.
It leverages a position-aware fusion mechanism of rally progress and player styles and incorporates encoder-decoder extractors to model the spatio-temporal dynamics of the game. The Transformer-based Rally Extractor (TRE) captures the global temporal evolution of the rally, while the Transformer-based Player Extractor (TPE) explicitly disentangles and processes information specific to each player. A key innovation of ShuttleNet is the Position-aware Gated Fusion Network (PGFN), which adaptively integrates rally-level context by jointly considering contextual dependencies and positional significance.
In addition, the integration of badminton video data with emerging technologies, such as wearable sensors \cite{seong2024counterfactualexplanationbasedbadmintonmotion}, virtual reality (VR) \cite{VR_badminton}, augmented reality (AR) \cite{chien2023automatedhitframedetectionbadminton}, and large language models (LLMs) \cite{chiang2024badgebadmintonreportgeneration} has attracted increasing attention and is being explored.

\section{Fine-Badminton}
\label{sec:fine-Badminton}
To address the lack of professional badminton benchmarks in temporal action localization research, we describe the construction of Fine-Badminton, a domain-specific dataset designed for fine-grained temporal kinematic analysis.

\subsection{Dataset Generation}
\label{subsec:dataset}
\emph{Raw data collection}: The raw videos are collected from major international tournaments (\emph{e.g.}, Sudirman Cup and World Championships, 2011--2012) and amateur competitions. All videos are downloaded from YouTube and resampled to 25 fps. We retain broadcast-view footage (100\%) at 720p resolution and restrict the dataset to professional singles matches to ensure consistent motion patterns and reduce variability from doubles-specific tactics.

\emph{Annotation process}: The annotation process follows a multi-stage framework to ensure temporal accuracy and label consistency. We define 29 fine-grained badminton action categories through consultation with two professional athletes, providing consistent terminology for annotators. Five annotators with sports science backgrounds undergo a 10-hour training protocol covering annotation guidelines and edge-case handling, followed by a qualification test ($\geq$95\% accuracy on 500 reference frames) before formal annotation. A custom-developed tool is used to annotate sequential videos, enabling precise labeling of action categories and accurate marking of action start and end timestamps.

\emph{Dataset statistic}: The final dataset consists of 31 professionally recorded badminton singles matches, containing 2948689 frames at 25 fps (approximately 32.76 hours).

In the experimental setup, videos are segmented into game-level clips by detecting inactive periods of more than 150 consecutive frames without active play to identify round boundaries, reducing redundant content and retaining informative action sequences.

\begin{figure*}[!t]
    \centering
    \subfloat[]{\includegraphics[width=0.32\linewidth, trim=0 0 0 20, clip]{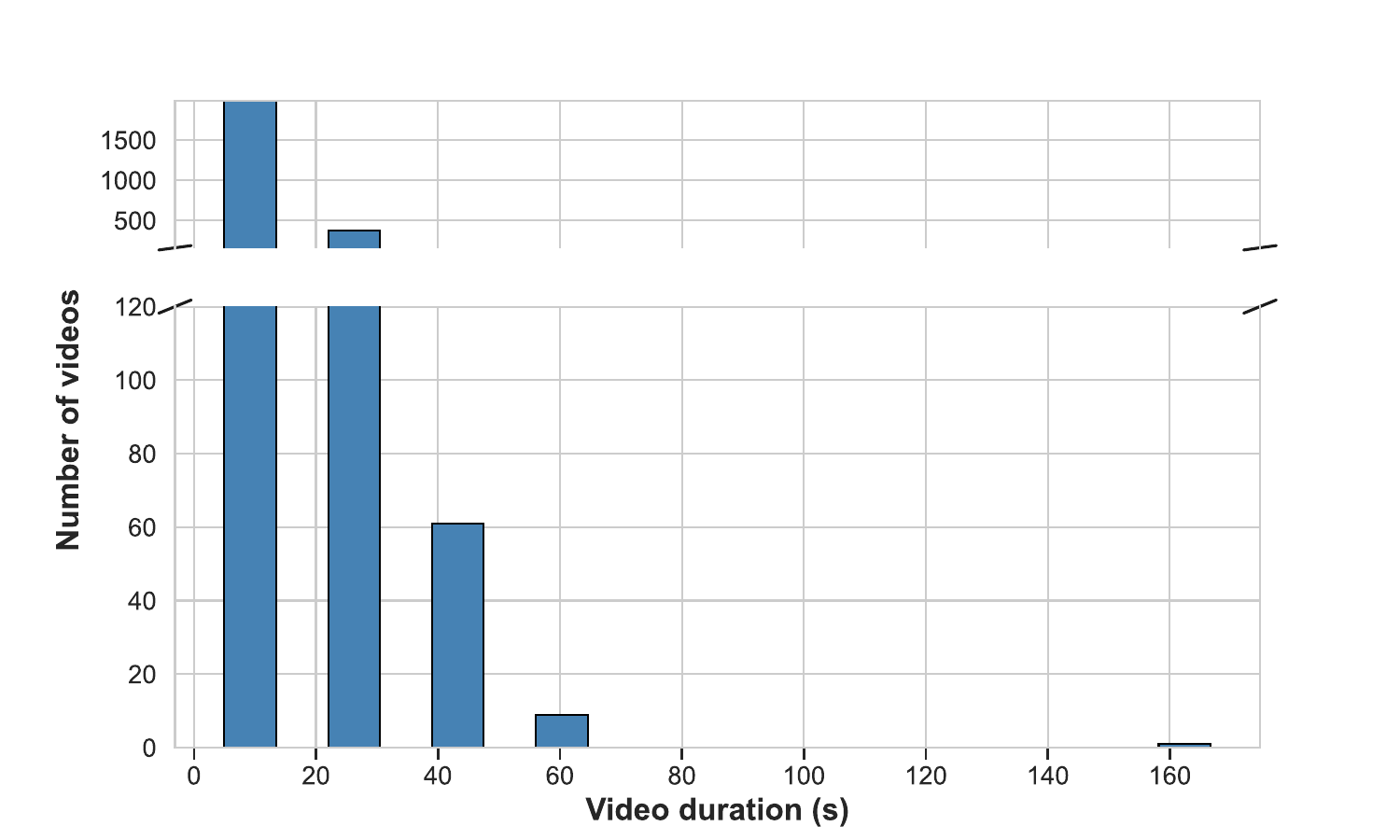}\label{fig:video_duration}}%
    \hfil
    \subfloat[]{\includegraphics[width=0.32\linewidth, trim=0 0 0 20, clip]{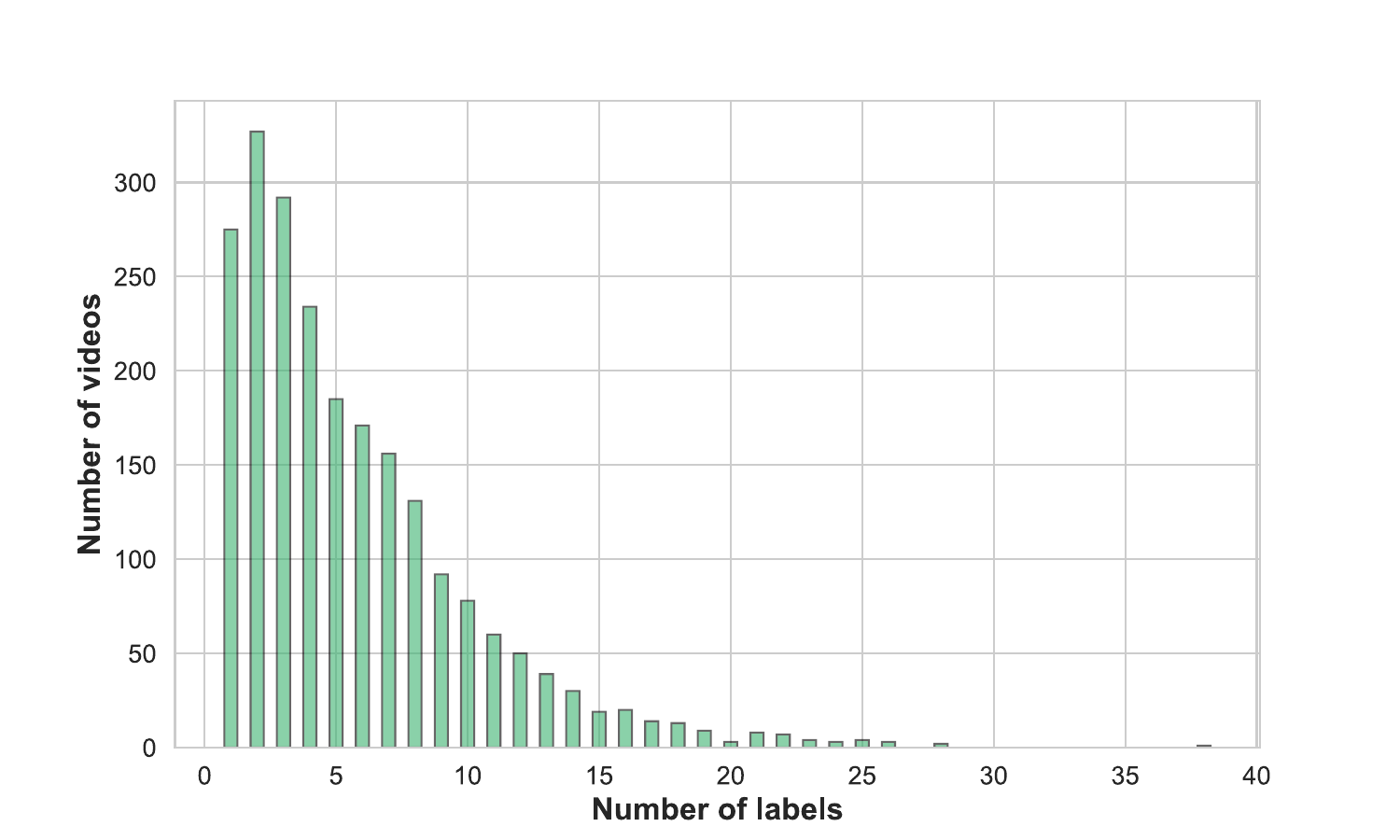}\label{fig:label_counts}}%
    \hfil
    \subfloat[]{\includegraphics[width=0.32\linewidth, trim=0 0 0 20, clip]{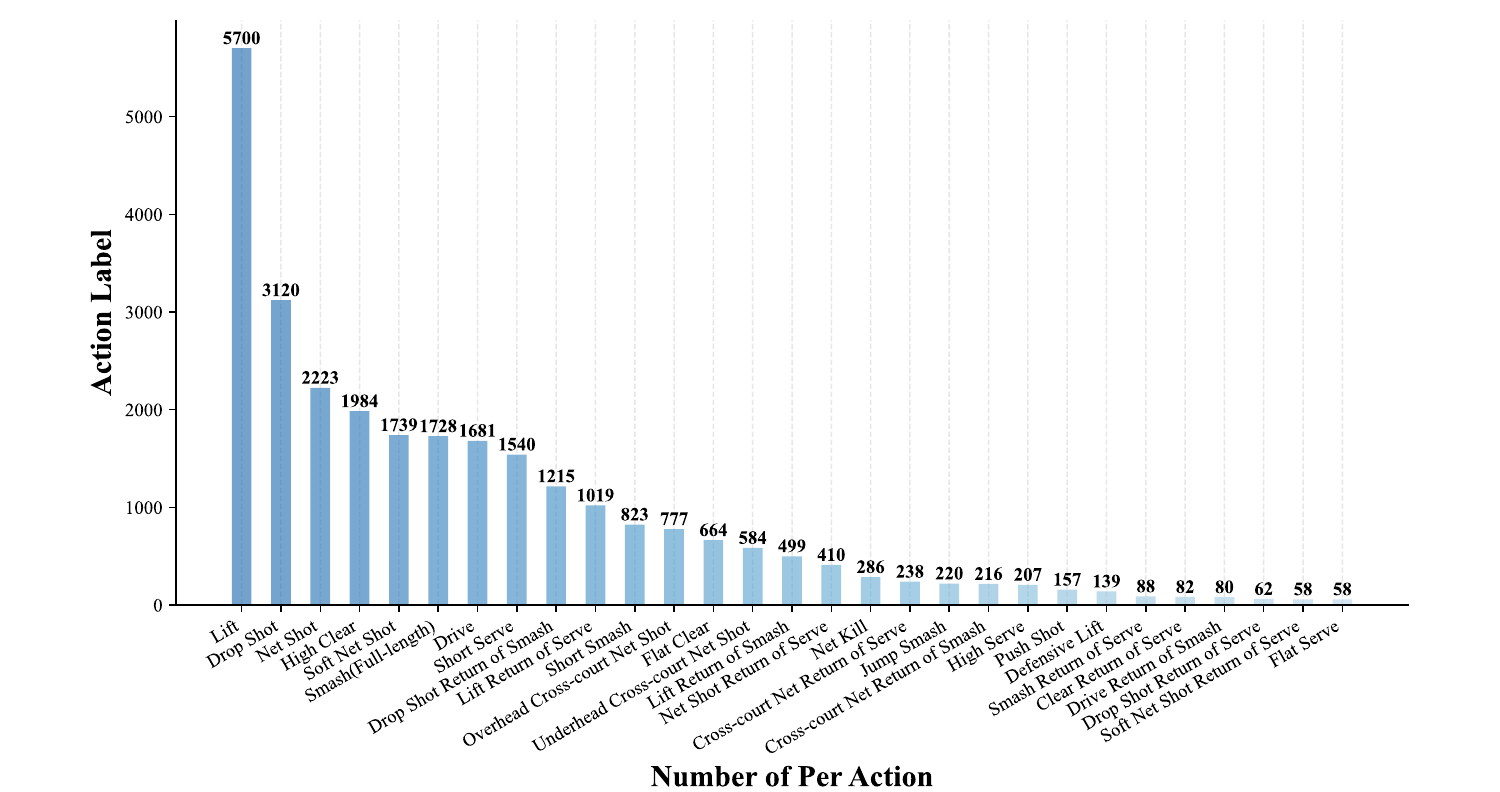}\label{fig:dataset_labels}}%
    \caption{Dataset statistics. (a) Video duration distribution (broken y-axis). (b) Distribution of label counts per video. (c) Distribution of action instances per class.}
    \label{fig:dataset_statistics}
\end{figure*}

Fig.~\ref{fig:dataset_statistics} illustrates three key characteristics. Fig.~\ref{fig:video_duration} illustrates the distribution of segmented clip durations, where the majority of segments fall within the 0 – 60 s interval. The broken y-axis highlights low-frequency long-duration clips without obscuring shorter segments. Fig.~\ref{fig:label_counts} presents the distribution of action label counts per segmented clip. Most segments contain fewer than five labels, indicating that individual rounds typically involve a limited number of distinct actions.

\begin{figure}[!t]
  \centering
   \includegraphics[width=0.8\linewidth]{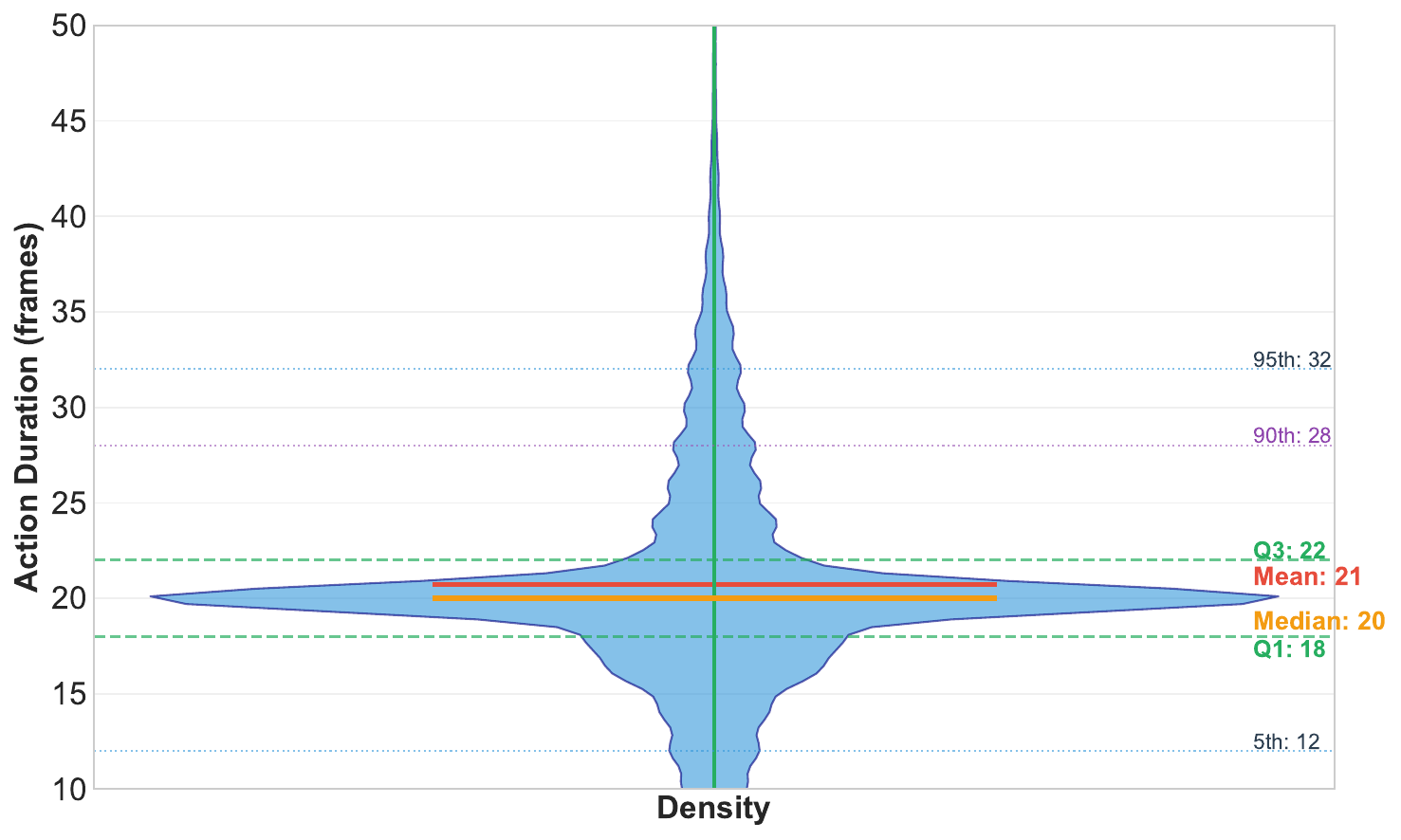}
   \caption{Distribution of action durations across all annotations in Fine-Badminton.}
   \label{fig:duration_violin}
\end{figure}

Through frame-level temporal annotation, we obtain 27597 action instances spanning 29 fine-grained stroke categories, including "Short Smash", "High Clear", and "Drop Shot". Most actions last 10 - 50 frames (0.4 - 2 s) (Fig.~\ref{fig:duration_violin}).
The 29 action categories exhibit a long-tailed distribution (Fig.~\ref{fig:dataset_labels}, 27597 instances in total). Based on the deviation $\sigma$ of per-category action instance counts, actions are grouped into four categories: 

\begin{enumerate}
    \item High-frequency actions: $\sigma \geq 0.7$
    \item Intermediate-frequency actions: $0 \leq \sigma < 0.7$
    \item Low-frequency actions: $-0.7 \leq \sigma < 0$
    \item Rare actions: $\sigma < -0.7$
\end{enumerate}

High-frequency actions range from "Lift" (5,700 instances, 20.65\%) to "High Clear" (1,984 instances, 7.19\%). Intermediate-frequency actions range from "Soft Net Shot" (1,739 instances, 6.30\%) to "Lift Return of Serve" (1,019 instances, 3.69\%). Low-frequency actions range from "Short Smash" (823 instances, 2.98\%) to "Defensive Lift" (139 instances, 0.50\%). Rare actions range from "Smash Return of Serve" (88 instances, 0.32\%) to "Flat Serve" (58 instances, 0.21\%).

Fig.~\ref{fig:duration_violin} shows the distribution of annotated action durations in Fine-Badminton. Most actions are concentrated around 18 frames, reflecting the rapid execution of badminton strokes. Actions longer than 22 frames exhibit greater variability and typically correspond to strokes with longer shuttlecock trajectories, such as "High Clear" and "Flat Clear".

\subsection{Quality Control}
\label{subsec:quality_control}

We adopt a three-stage annotation protocol to ensure dataset quality. First, five trained annotators independently label the entire dataset following unified guidelines. Second, three senior annotators review a randomly selected 20\% subset, and inconsistencies are resolved via majority voting. Third, two professional badminton coaches audit 30\% of ambiguous cases (e.g., unclear stroke transitions), leading to a final inter-review agreement exceeding 98\%.

Compared with existing datasets such as ShuttleSet\cite{ShuttleSet}, Badminton Olympic\cite{badminton-dataset-towards-olympic}, BadmintonDB\cite{BadmintonDB}, and VideoBadminton\cite{VideoBadminton}, we provide interval-based annotations that explicitly define both action onset and offset, together with a finer-grained taxonomy of 29 stroke categories. In addition, Fine-Badminton contains a larger number of annotated action segments, providing more comprehensive coverage of professional badminton matches and more detailed temporal supervision. A representative subset of ten annotated videos is released on Zenodo \cite{fine-badminton2026}.

\section{Methods}
\label{sec:methods}

In this section, we introduce the notation and symbolic representations. Next, we present the overall architecture of the TAL model. Finally, we detail a spatio-temporally enhanced adapter module.

\subsection{Notations and Definitions}

Temporal action localization can be defined as follows: given an untrimmed video $\mathbf{X} \in \mathbb{R}^{C\times H\times W\times T}$, where $H$ and $W$ denote the spatial height and width of each frame, $T$ denotes the total number of frames, and $C$ denotes the number of input channels. Given the temporal action annotations of this video, the ground-truth annotation set is represented as $\mathbf{X}_{gt}=\{\mathbf{x}_{gt}=(t_s,t_e,c)\}_{i=1}^N$, where $t_s$ and $t_e$ respectively denote the start time and end time of the corresponding action segment, and $c$ denotes the action category label of $\mathbf{X}_{gt}$. Here, $N$ represents the number of ground-truth action instances in video $\mathbf{X}$. The objective of TAL is to predict a candidate proposal set $\mathbf{X}_p=\{\mathbf{x}_p=(\hat{t}_s,\hat{t}_e,\hat{c},s)\}_{i=1}^{N_p}$ that accurately matches $\mathbf{X}_{gt}$, where $s$ denotes the predicted confidence score.

\subsection{Overview}
\label{subsec:overview}

The foundation of our proposed approach builds upon AdaTAD \cite{adaTad}, an end-to-end deep learning framework that has demonstrated strong performance on benchmark datasets such as ActivityNet and THUMOS14.

Our research aims to enhance the framework with a novel adapter module. This enhancement is designed to improve the model's ability to detect and localize badminton strokes.

Here's an overview of the model's structure:

\emph{Backbone}: 
We adopt VideoMAE-B \cite{videomae} as the backbone network. VideoMAE is a self-supervised video representation learning framework that learns spatio-temporal representations through masked autoencoding. This approach eliminates the need for external annotations by training the model to reconstruct masked video tokens, thereby enabling effective feature learning from raw video data. VideoMAE captures rich spatio-temporal characteristics of videos, which are essential for action understanding and temporal localization. In our framework, VideoMAE serves as the feature extractor, producing high-level spatio-temporal representations from the input video, which are subsequently fed into the adapter module and the detection head.

\emph{Adapter}: 
A key enhancement to the model is the introduction of the Decoupling Spatio-Temporal Adapter (DSTA). In DSTA, a depth-wise convolution operation and three convolutional branches are used to separately model temporal and spatial features. This module is integrated into the backbone network to facilitate task-specific adaptation for fine-grained badminton stroke localization. It represents the core methodological contribution of our work, enabling the model to capture badminton-specific spatio-temporal characteristics with minimal increase in trainable parameters.

\emph{Head}:
We use ActionFormer \cite{actionFormer} as the action detection head, which is a transformer-based detector that is robust across benchmark datasets. This head is responsible for action classification and temporal boundary localization after spatio-temporal feature extraction, and constitutes a crucial component for end-to-end temporal action detection.

\emph{Loss}:
We adopt focal loss with sigmoid activation. Focal loss is designed to address class imbalance. By down-weighting easy samples, focal loss emphasizes hard examples during training, thereby mitigating the imbalance between positive and negative samples.

\subsection{DSTA}
\label{sec:DSTA}

We introduce refinements to the TIA module to optimize its performance.
While the original design is effective, it exhibits limitations in modeling fine-grained temporal dynamics, particularly in scenarios requiring precise temporal boundary discrimination. Accordingly, our modifications focus on enhancing the feature extraction capability of the TIA module.

Specifically, we revise the depth-wise convolution (DWConv) component. In the original design, DWConv is employed to model temporal information, whereas the subsequent feed-forward down-project operates solely along the channel dimension, without explicitly considering the spatial dimensions of the feature maps. This design restricts the joint modeling of spatio-temporal correlations. To address this limitation, we introduce the following modifications.

\begin{figure}[!t]
  \centering
   \includegraphics[width=0.8\linewidth, trim=0 0 0 10, clip]{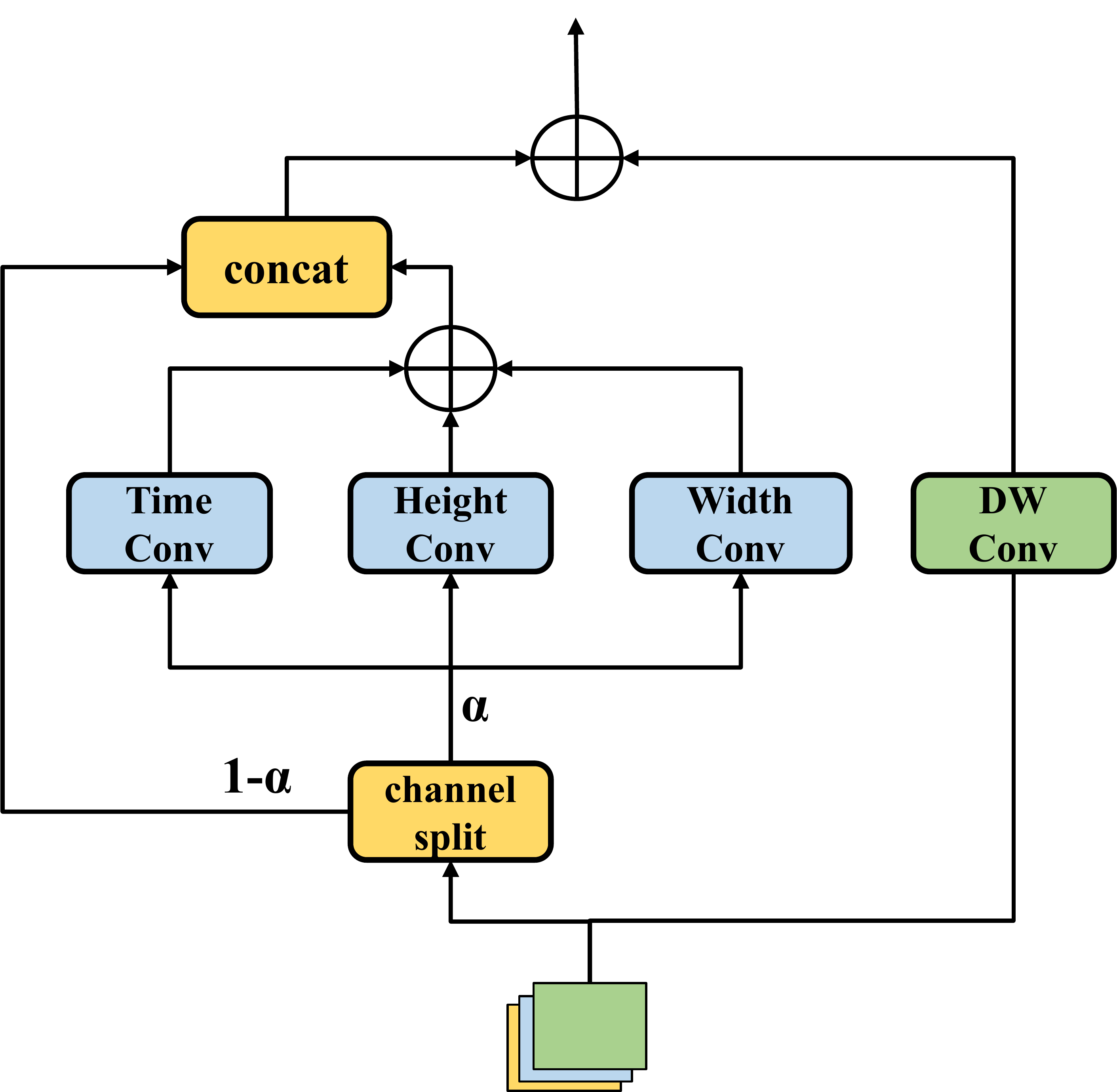}
   \caption{Architecture of the proposed DST module within the DSTA.}
   \label{fig:DST module}
\end{figure}

As illustrated in Fig.~\ref{fig:DST module}, the first modification replaces the original single DWConv with two parallel processing branches. The first branch preserves the original DWConv structure, thereby retaining its ability to capture local temporal patterns and short-term dependencies.

The second stream introduces a tri-branch convolutional layer consisting of three convolution kernels:
\begin{itemize}
    \item $3 \times1\times1$ (temporal dimension),
    \item $1\times3\times1$ (height dimension),
    \item $1\times1\times3$ (width dimension).
\end{itemize}
These kernels explicitly model directional dependencies along temporal and spatial axes, enabling effective spatio-temporal feature aggregation. Such a design enhances the model's capacity to encode subtle motion details, which is particularly beneficial for accurately characterizing rapid movements. From the perspective of the entire DSTA in Fig.~\ref{fig:DSTA}, the DSTA module can be formally expressed by Eq.~\eqref{eq:1}. Note that $\mathbf{x}$ and $\mathbf{x}^\prime$ denote the input and output feature tensors, respectively, sharing the same dimensionality $\mathbb{R}^{C\times T\times H\times W}$. The DSTA includes a channel-wise down-projection fully connected (FC) layer with parameter $\mathbf{W}_{\text{down}} \in \mathbb{R}^{C\times N}$, where $N$ denotes the intermediate embedding dimension, satisfying $0 < N < C$. The projected features are then passed through a non-linear activation function $\boldsymbol{\sigma}$. Subsequently, the features are processed by the proposed DST module, which will be detailed later. The resulting features are then fed into an intermediate FC transformation $\mathbf{W}_{\text{mid}}\in\mathbb{R}^{N\times N}$ and combined with the activated features via a residual addition operation, followed by a channel-wise up-projection layer $\mathbf{W}_{\text{up}}\in\mathbb{R}^{N\times C}$. Finally, the output feature magnitude is modulated by a learnable scaling parameter $\beta$.

\begin{subequations}
\label{eq:1}
\begin{align}
\bar{\mathbf{x}}\hphantom{^\prime} &= \sigma(\mathbf{W}^\top_{\text{down}}\mathbf{x}) \label{eq:1A}\\
\hat{\mathbf{x}}\hphantom{^\prime} &= \mathbf{W}^\top_{\text{mid}}\mathbf{DST}(\bar{\mathbf{x}}) + \bar{\mathbf{x}} \label{eq:1B}\\
\mathbf{x}^\prime &= \beta\cdot\mathbf{W}_{\text{up}}^\top\cdot\hat{\mathbf{x}}+\mathbf{x} \label{eq:1C}
\end{align}
\end{subequations}

Next, we describe the detailed implementation of the proposed DST module. The first stream preserves the original DWConv structure to ensure stable extraction of local temporal features. The input channels are split according to a ratio of $\alpha$, where $\alpha C$ channels are routed to the spatial-temporal branches, while the remaining $(1-\alpha)C$ channels are retained as identity features. By adjusting $\alpha$, the model can explicitly control the relative emphasis between spatial and temporal feature modeling.
The second stream consists of three parallel convolutional branches operating on $\alpha C$ channels. Specifically, the $3\times1\times1$ convolution captures temporal dependencies, the $1\times3\times1$ convolution models spatial variations along the height dimension, and the $1\times1\times3$ convolution captures spatial variations along the width dimension. Through this design, temporal dynamics and spatial contextual information are explicitly decoupled, enabling the adapter to learn richer and more discriminative spatio-temporal representations. Since badminton actions exhibit highly coupled yet heterogeneous spatio-temporal characteristics, this decoupled structure improves the model’s generalization capability by separately modeling motion patterns and spatial positional cues, thereby enhancing adaptability across diverse gameplay scenarios.
The feature maps generated by the three convolutional branches are fused through element-wise addition to obtain a unified spatio-temporal representation. Subsequently, the remaining $(1-\alpha)C$ channels are concatenated with the fused representation. The resulting feature tensors are denoted as $\bar{\mathbf{X}}_1 \in \mathbb{R}^{\alpha C\times T\times H\times W}$ and $\bar{\mathbf{X}}_2 \in \mathbb{R}^{(1-\alpha) C\times T\times H\times W}$, respectively. Here, $\hat{\mathbf{x}}1$ denotes the concatenation of $\hat{\mathbf{x}}{11}$ and $\bar{\mathbf{x}}_2$, while $\hat{\mathbf{x}}$ represents the final output of the DSTA module. The overall processing pipeline of the DST module is formally expressed in Eq.~\eqref{eq:2}.

\begin{subequations}
    \label{eq:2}
    \begin{align}
        \hat{\mathbf{x}}_{11} &=\ \mathbf{Conv}_{3\times1\times1}(\bar{\mathbf{x}}_1)+\mathbf{Conv}_{1\times3\times1}(\bar{\mathbf{x}}_1)\nonumber \\
        &\hphantom{=\ \ }+\mathbf{Conv}_{1\times1\times3}(\bar{\mathbf{x}}_1) \label{eq:2A}\\
        \hat{\mathbf{x}}_1\hphantom{_1} &= [\hat{\mathbf{x}}_{11};\bar{\mathbf{x}}_2] \label{eq:2B}\\
        \hat{\mathbf{x}}_2\hphantom{_1} &=\ \mathbf{DWConv}(\bar{\mathbf{x}}) \label{eq:2C}\\
        \hat{\mathbf{x}}\hphantom{_11} &=\ \hat{\mathbf{x}}_1+\hat{\mathbf{x}}_2 \label{eq:2D}
    \end{align}
\end{subequations}

DWConv in the first stream is implemented as a 3D depth-wise convolution with a kernel size $(k,1,1)$ along the temporal dimension and a group size of $m$. The output of the second stream is element-wise added to the DWConv-processed output of the first stream, thereby producing the final DSTA output feature representation.

\section{Experiments}
\label{sec:experiments}

In this section, we describe the experimental protocol and implementation settings, including the datasets, evaluation metrics, and hyperparameter configurations used in our experiments. Subsequently, we compare the proposed method with the baseline model. Finally, we conduct ablation studies on the proposed DST module and provide a detailed analysis of its effectiveness.

\subsection{Datasets and Metrics}
\label{subsec:datasets and metrics}

The ShuttleSet dataset \cite{ShuttleSet} and the Fine-Badminton dataset, introduced in Section~\ref{subsec:badminton_video_dataset} and Section~\ref{sec:fine-Badminton}, are used for training. To adapt ShuttleSet to the TAL setting, we convert each stroke-point annotation into a temporal interval due to the absence of explicit boundaries. Specifically, for a stroke at frame $t$, we construct a fixed symmetric window $[t-9, t+9]$ (19 frames).
This provides a deterministic approximation of stroke duration suitable for short badminton motions. All intervals are clipped to valid frame ranges, and each annotation is represented as $(\text{start}, \text{end}, \text{label})$. All intervals are clipped to valid frame ranges and manually inspected to remove clearly invalid or degenerate segments.

Following common practice, we adopt mean Average Precision (mAP) under multiple temporal Intersection-over-Union (tIoU) thresholds chosen from \{0.3, 0.4, 0.5, 0.6, 0.7\}, and the average mAP over these thresholds is also reported for comprehensive performance assessment.

\subsection{Implementation Details}

The proposed method is fully described to allow straightforward implementation. We implement our method using PyTorch 2.0.1, MMCV 2.0.1 \cite{mmcv}, and MMAction2 1.1.0 \cite{2020mmaction2} on two independent machines, each equipped with an NVIDIA RTX 4090 GPU. Activation checkpointing is employed to reduce GPU memory consumption during training. As introduced in Section~\ref{subsec:overview}, we adopt ActionFormer \cite{actionFormer} as the action detection head. The learning rate of the adapter modules inserted into the backbone is selected from $\{1\times10^{-3}, 5\times10^{-4}, 1\times10^{-4}, 5\times10^{-5}, 1\times10^{-5}\}$, while all other backbone parameters are frozen to ensure parameter-efficient adaptation. Unless otherwise specified, the channel split ratio $\alpha$ is set to 0.5 in all experiments, as it achieves the best trade-off between preserving original features and enhancing spatio-temporal modeling, preventing either branch from dominating the learned representations.

First, we resample all videos to a unified frame rate of 25 fps, and adjust the corresponding annotation timestamps accordingly. To reduce interference from background content and non-action-related visual information, we segment videos by identifying intervals where the gap between the end frame of the previous action and the start frame of the subsequent action exceeds 150 frames, and extract clips accordingly. We additionally retain 10 frames before and after each segmented clip to preserve temporal context.
The segmentation threshold of 150 frames (approximately 6 seconds) is empirically chosen, since intervals longer than this can reasonably be regarded as breaks between rally rounds in professional badminton matches. Meanwhile, it is not set too small to avoid over-segmentation caused by short broadcast operations such as camera viewpoint switching or replay insertions, which frequently occur in real match videos.
For both ShuttleSet and Fine-Badminton, we randomly sample a temporal window of 768 frames with a stride of 4 as model input. After passing through the video encoder, the extracted features are temporally resized to a fixed length of 768. The spatial resolution of input frames is uniformly set to $160 \times 160$.

\begin{table*}[!t]
  \centering
  \caption{Results on Fine-Badminton and processed ShuttleSet, measured by mAP(\%) at different tIoU thresholds. We report our best results in \textbf{bold}.}
  \begin{tabular}{l|c|rrrrr|>{\columncolor{lightgray}}c}
    \hline
    Dataset & Adapter & mAP@0.3 & mAP@0.4 & mAP@0.5 & mAP@0.6 & mAP@0.7 & Avg. \\
    \hline
    \multirow{7}{*}{ShuttleSet} &TIA\cite{adaTad}(baseline) & 72.26\% & 72.16\% & 72.00\% & 71.46\% & 68.85\% & 71.35\% \\ 
        &3D-SENet\cite{ada_improve_3} & 71.41\% & 71.23\% & 71.03\% & 70.11\% & 67.50\% & 70.26\%  \\ 
        & AIM\cite{yang2023aim} & 73.78\% & 73.74\% & 72.63\% & 72.99\% & 71.72\% & 73.17\%\\
        & ST-Adapter\cite{pan2022st-adapter} & 14.57\% & 13.81\% & 12.96\% & 12.22\% & 11.32\% & 12.98\%\\
        & LoRA\cite{hu2021loralowrankadaptationlargeLoRA} & 60.99\% & 60.62\% & 60.05\% & 57.89\% & 52.33\% & 58.38\%\\
        &EVL\cite{lin2022frozenEVL} & 15.51\% & 14.68\% & 13.96\% & 13.06\% & 11.85\% & 13.81\%\\
        &\textbf{DSTA} & \textbf{75.17}\% & \textbf{75.10}\% & \textbf{75.04}\% & \textbf{74.84}\% & \textbf{73.18}\% & \textbf{74.67}\%\\ 
    \hline
    \multirow{7}{*}{Fine-Badminton} & TIA(baseline) & 66.68\% & 66.14\% & 65.64\% & 63.10\% & 58.81\% & 64.08\%  \\ 
        & 3D-SENet & 61.40\% & 61.15\% & 60.23\% & 58.08\% & 54.52\% & 59.08\%  \\
        & AIM & 63.78\% & 63.51\% & 62.40\% & 60.54\% & 56.71\% & 61.39\%\\
        & ST-Adapter & 10.05\% & 9.13\% & 7.37\% & 5.68\% & 4.14\% & 7.27\%\\
        & LoRA & 45.50\% & 44.40\% & 42.06\% & 38.30\% & 31.31\% & 40.31\%\\
        &EVL & 10.75\% & 9.90\% & 8.53\% & 6.92\% & 5.45\% & 10.75\%\\
        & \textbf{DSTA} & \textbf{68.47}\% & \textbf{67.83}\% & \textbf{67.52}\% & \textbf{65.88}\% & \textbf{61.44}\% & \textbf{66.23}\%  \\
    \hline
  \end{tabular}
  \label{tab:comparison}
\end{table*}

\subsection{Parameter and Computational Efficiency}

In addition to performance improvements, we further analyze the parameter efficiency and computational cost of our proposed DSTA. Since our method follows an adapter-based paradigm, it is crucial to verify whether the observed performance gains come at the expense of increased model complexity.

As shown in Table~\ref{tab:efficiency}, all methods are evaluated under a unified setting, where the FLOPs are computed with a temporal input window of 768 frames and a chunk size of 48, ensuring a fair and consistent comparison across different approaches. For our method, the channel split ratio $\alpha$ in the DST module is set to 0.5. This unified evaluation protocol eliminates discrepancies caused by different input configurations, allowing a more reliable comparison of computational efficiency. Table~\ref{tab:efficiency} compares our method with full fine-tuning and several representative adapter-based approaches in terms of total parameters, trainable parameters, computational cost (FLOPs), and detection performance. Specifically, full fine-tuning of AdaTAD requires updating all parameters of the backbone network, resulting in a significantly larger number of trainable parameters (86.227 M). In contrast, lightweight adapter-based methods such as TIA and 3D-SENet drastically reduce the number of trainable parameters (1.335 M and 3.664 M, respectively) while incurring moderate FLOPs (428.17 and 410.56 GFLOPs). LoRA, although extremely parameter-efficient with only 0.229 M trainable parameters, exhibits a notably higher FLOPs (1136.73 GFLOPs) due to additional low-rank projections applied over long temporal sequences. More complex adapter designs, including ST-Adapter, Adapt Image Models (AIM)\cite{yang2023aim}, and EVL, introduce additional spatial-temporal branches, increasing both trainable parameters and FLOPs (ranging from 2.886 M to 13.308 M and from 955.51 to 2812.58 GFLOPs).

Our proposed DSTA strikes a favorable balance between parameter efficiency and computational cost. Despite incorporating additional spatial branches for enhanced feature modeling, DSTA introduces only a marginal increase in trainable parameters (4.010 M - 4.017 M) while maintaining FLOPs comparable to simpler adapter designs (~428 GFLOPs). This indicates that the performance improvements are achieved through effective decoupling of spatial and temporal representations, rather than merely enlarging model capacity. Consequently, DSTA consistently improves detection accuracy on both ShuttleSet and Fine-Badminton datasets (see Table~\ref{tab:comparison}) while keeping the computational overhead low, making it particularly suitable for practical scenarios with limited memory and compute resources.

Overall, the results demonstrate that DSTA achieves a strong balance between efficiency and accuracy, validating its effectiveness as a lightweight yet powerful adaptation module for fine-grained temporal action localization.

\begin{table}[!t]
    \centering
    \caption{Comparison of parameter efficiency and computational cost.}
    \begin{tabular}{lcc}
        \hline
        Method & Trainable Params (M) & FLOPs (G) \\
        \hline
        Full Fine-tuning (AdaTAD) & 86.227 & 274.62 \\
        3D-SENet & 3.664 & 410.56 \\
        LoRA & 0.229 & 1136.73 \\
        ST-Adapter & 2.886 & 2812.58 \\
        AIM & 10.651 & 955.51 \\
        EVL & 13.308 & 1379.45 \\
        \hline
        TIA (Baseline) & 1.335 & 428.17 \\
        + Conv T & 4.010 & 427.90 \\
        + Conv TH & 4.014 & 428.04 \\
        + Conv THW (DSTA) & 4.017 & 428.17 \\
        \hline
    \end{tabular}
  \label{tab:efficiency}
\end{table}

\subsection{Comparison with SoTA Methods}

We present the results of our proposed DSTA adapter compared to TIA \cite{adaTad} and 3D-SENet \cite{ada_improve_3} on ShuttleSet \cite{ShuttleSet} and Fine-Badminton in Table~\ref{tab:comparison}. All adapters are evaluated under the same experimental protocol using the evaluation metrics illustrated in Section~\ref{subsec:datasets and metrics}.

\emph{Results on ShuttleSet}: ShuttleSet \cite{ShuttleSet} is an expert-annotated, rally-level badminton singles dataset originally designed for tactical analysis. Following the preprocessing strategy described in Section~\ref{subsec:datasets and metrics}, we adapt ShuttleSet to the TAL setting used in this study.

As shown in Table~\ref{tab:comparison}, on the processed ShuttleSet dataset, our proposed DSTA consistently achieves the best performance across all tIoU thresholds. Specifically, DSTA improves the average mAP from 71.35\% to 74.67\% compared to the TIA-based baseline~\cite{adaTad}, yielding a gain of 3.32\%. In comparison with AIM~\cite{yang2023aim}, which achieves an average mAP of 73.17\%, DSTA further improves performance by 1.5\%. Moreover, DSTA outperforms other adapter-based methods such as 3D-SENet~\cite{ada_improve_3} (70.26\%) by a significant margin of 4.41\%.
In addition to overall performance, DSTA shows consistent improvements across all evaluation thresholds. For example, at tIoU=0.5, the mAP increases from 72.00\% (TIA) to 75.04\%, while at the stricter tIoU=0.7, the improvement is more pronounced, rising from 68.85\% to 73.18\%. These results indicate that the proposed method not only improves detection accuracy but also enhances temporal boundary localization under more stringent evaluation criteria.

Overall, the consistent gains across different thresholds and baselines demonstrate the effectiveness of DSTA in improving temporal action localization performance on ShuttleSet.

\begin{table*}[!t]
    \centering
    \caption{Ablation of different adapter architectural designs on ShuttleSet and Fine-Badminton, measured by mAP(\%) at different tIoU thresholds. VideoMAE-B is used to conduct the following experiments. We report our best results in \textbf{bold}.}
    \begin{tabular}{l|l|rrrrr|>{\columncolor{lightgray}}c}
        \hline
        Dataset & Adapter & mAP@0.3 & mAP@0.4 & mAP@0.5 & mAP@0.6 & mAP@0.7 & Avg \\
        \hline
        \multirow{4}{*}{ShuttleSet} & Baseline(TIA) & 72.26\% & 72.16\% & 72.00\% & 71.46\% & 68.85\% & 71.35\%  \\ 
         & +Conv T & 73.28\% & 73.23\% & 73.13\% & 72.56\% & 70.18\% & 72.48\%  \\ 
         & +Conv TH & 73.65\% & 73.56\% & 73.47\% & 73.10\% & 71.20\% & 73.00\%  \\ 
         & +Conv THW(DSTA) & \textbf{75.17}\% & \textbf{75.10}\% & \textbf{75.04}\% & \textbf{74.84}\% & \textbf{73.18}\% & \textbf{74.67}\%\\ 
        \hline
        \multirow{4}{*}{Fine-Badminton} & Baseline(TIA) & 66.68\% & 66.14\% & 65.64\% & 63.10\% & 58.81\% & 64.08\%  \\ 
         & +Conv T & 66.60\% & 66.30\% & 65.88\% & 64.36\% & 59.61\% & 64.55\%  \\ 
         & +Conv TH & 67.25\% & 66.80\% & 66.25\% & 64.45\% & 59.30\% & 64.81\%  \\
         & +Conv THW(DSTA) & \textbf{68.47}\% & \textbf{67.83}\% & \textbf{67.52}\% & \textbf{65.88}\% & \textbf{61.44}\% & \textbf{66.23}\%  \\
         \hline
    \end{tabular}
  \label{tab:ablation}
\end{table*}

\emph{Results on Fine-Badminton}: On our Fine-Badminton dataset, our proposed DSTA consistently improves performance over all compared methods. Specifically, DSTA achieves an average mAP of 66.23\%, outperforming TIA \cite{adaTad} by a clear margin (64.08\% $\rightarrow$ 66.23\%). Compared with AIM \cite{yang2023aim}, the improvement is even more significant, with performance increasing from 61.39\% to 66.23\%. In addition, DSTA also demonstrates a notable advantage over 3D-SENet \cite{ada_improve_3}, improving from 59.08\% to 66.23\%. These results indicate that the proposed decoupled spatio-temporal modeling is more effective than existing adapter-based designs for fine-grained action localization.

The performance gap is more pronounced on Fine-Badminton than on ShuttleSet, which can be attributed to the increased difficulty of the dataset. As introduced in Section~\ref{subsec:dataset}, Fine-Badminton contains 29 fine-grained action categories, significantly more than the 18 categories in ShuttleSet \cite{ShuttleSet}. The higher label granularity introduces stronger inter-class similarity and finer temporal distinctions, making it more challenging to accurately localize and classify actions. Consequently, all compared methods exhibit lower absolute performance on Fine-Badminton, reflecting the intrinsic difficulty of the task.

Despite this challenge, DSTA maintains consistent improvements across all baselines, demonstrating its effectiveness in capturing subtle spatio-temporal patterns. The advantage of DSTA can be attributed to its decoupled design, which explicitly models temporal dynamics and spatial variations through dedicated branches. This enables the model to better distinguish fine-grained stroke differences and to more precisely localize short-duration actions.


Overall, these results demonstrate that DSTA not only improves detection accuracy but also exhibits strong robustness under more challenging fine-grained settings. This highlights its superior generalization capability and effectiveness for real-world badminton video analysis, where actions are rapid, subtle, and highly similar across categories.

\emph{Performance under different tIoU thresholds}: 
To further evaluate the localization capability of our method, we analyze the performance under different tIoU thresholds, where tIoU $\in \{0.3, 0.4, 0.5, 0.6, 0.7\}$.

As shown in the Table~\ref{tab:comparison}, our proposed DSTA consistently outperforms the baseline TIA across all tIoU thresholds on both datasets.
Notably, the performance gap becomes more pronounced as the tIoU threshold increases (\emph{e.g.}, tIoU = 0.7), indicating that DSTA is more effective in capturing precise temporal boundaries.

This improvement can be attributed to the decoupling of spatio-temporal modeling design, which enhances the model’s ability to capture fine-grained motion patterns and temporal dynamics. The consistent gains across different thresholds demonstrate the robustness of our method under both coarse and strict localization settings.

\subsection{Internal Analysis}
\label{subsec:ablation and analysis}

In this section, we present systematic ablation studies and investigate the effect of the channel split ratio $\alpha$, aiming to evaluate the contribution of each component and analyze both overall performance and per-class AP variations.

\emph{Ablation analysis}:
As shown in Table~\ref{tab:ablation}, we evaluate three configurations: 1) baseline without DST, 2) partial DST with temporal and height-aware spatial branches, and 3) full DST with all branches. The results demonstrate consistent performance gains across components, validating the effectiveness of the decoupled design.

On ShuttleSet, introducing the temporal branch improves mAP from 71.35\% to 74.67\%, and removing any component leads to performance degradation, confirming the importance of each sub-module. DST achieves larger gains on the ShuttleSet dataset. Similar trends can also be observed on Fine-Badminton.

\begin{table}[!t]
\centering
\caption{Sensitivity analysis of channel split ratio $\alpha$ on Fine-Badminton. We report our best results in \textbf{bold}.}
\begin{tabular}{r|ccccc}
\hline
 $\alpha$ & 0.1 & 0.3 & 0.5 & 0.7 & 0.9 \\
\hline
mAP@0.3 & 67.10\% & 67.53\% & \textbf{68.47}\% & 67.38\% & 67.82\% \\
mAP@0.4 & 66.63\% & 67.23\% & \textbf{67.83}\% & 66.83\% & 67.43\% \\
mAP@0.5 & 66.31\% & 66.89\% & \textbf{67.52}\% & 66.27\% & 66.80\% \\
mAP@0.6 & 64.11\% & 65.06\% & \textbf{65.88}\% & 64.89\% & 65.33\% \\
mAP@0.7 & 59.21\% & 60.59\% & 61.44\% & 60.64\% & \textbf{61.46\%} \\
\hline
Avg. & 64.67\% & 65.46\% & \textbf{66.23}\% & 65.20\% & 65.77\% \\
\hline
\end{tabular}
\label{tab:alpha}
\end{table}


\emph{Sensitivity analysis}:
We analyze the sensitivity of our proposed method to the channel split ratio $\alpha$, which controls the proportion of channels allocated to the spatio-temporal branches.

To evaluate its impact, we conduct experiments on the Fine-Badminton dataset by varying $\alpha \in \{0.1, 0.3, 0.5, 0.7, 0.9\}$. The results are reported in Table~\ref{tab:alpha}.


From Table~\ref{tab:alpha}, the best performance is consistently achieved at $\alpha=0.5$, with an average mAP of 66.23\%, outperforming other configurations across all tIoU thresholds. This indicates that allocating equal modeling capacity between spatio-temporal transformation and residual feature preservation provides a more effective representation balance for fine-grained motion understanding.

Overall, the performance variation across different $\alpha$ values is relatively small (64.67\%–66.23\%), indicating that the proposed architecture is robust to channel partitioning. Notably, $\alpha=0.5$ consistently achieves the best performance, suggesting that a balanced feature allocation is beneficial for optimal spatio-temporal modeling.

\begin{figure}[!t]
    \centering
    \includegraphics[width=0.5\textwidth]{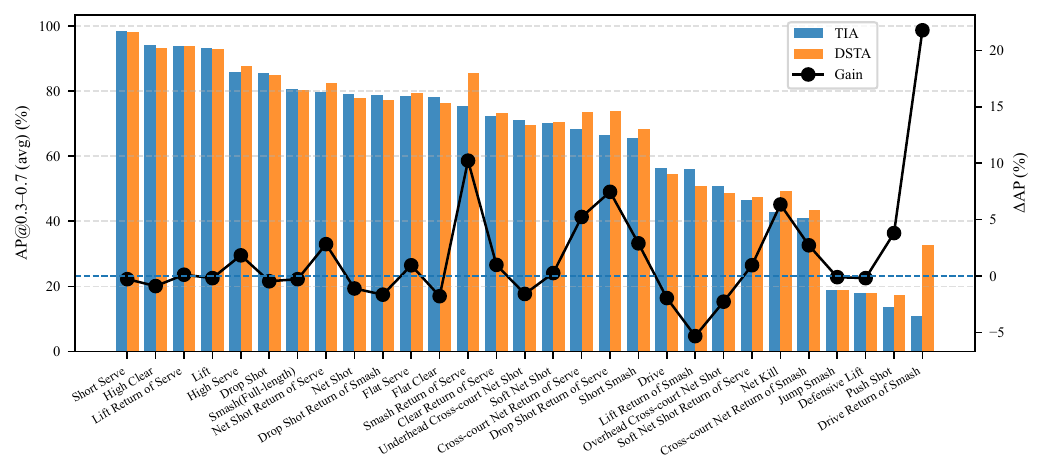}
    \caption{Per-class average precision (AP) comparison and AP gain of DSTA over the TIA baseline on Fine-Badminton. Action categories are sorted by the mean AP of the two methods. Bars indicate the AP of each method, and the black line represents the performance gain ($\Delta$AP).}
    \label{fig:per_class_ap}
\end{figure}

\emph{Per-class performance analysis}:
To further investigate how the proposed DSTA affects different action categories, we compare the per-class Average Precision (AP) of DSTA and the TIA baseline on Fine-Badminton, as shown in Fig.~\ref{fig:per_class_ap}. The action categories are sorted in descending order according to the AP achieved by the TIA baseline.

DSTA achieves substantial improvements in categories requiring precise modeling of rapid temporal dynamics and directional spatial cues. The most notable gain is observed for "Drive Return of Smash" (+21.79\%), which requires fast reaction and strong directional motion modeling. Significant improvements are also obtained for "Smash Return of Serve" and "Drop Shot Return of Serve".

In contrast, categories with relatively regular motion patterns, such as "Lift Return of Serve" and "Flat Serve", show only marginal changes, indicating that the baseline already captures their dominant temporal characteristics. A few categories exhibit moderate declines, with the largest decrease occurring for "Lift Return of Smash" (-5.33\%), which exhibits similar motion patterns to "Lift" and "Defensive Lift".

Overall, these results indicate that DSTA is more effective on fine-grained actions involving rapid reactions and directional motion, highlighting the benefit of explicitly decoupling temporal and spatial feature modeling.

\section{Conclusion}
\label{sec:conclusion}
In this work, we address fine-grained temporal action localization in professional badminton by introducing Fine-Badminton and the Decoupling Spatio-Temporal Adapter (DSTA). Fine-Badminton provides interval-based annotations for 29 fine-grained stroke categories across 31 professional matches, offering finer granularity than existing datasets such as ShuttleSet \cite{ShuttleSet} and Badminton Olympic \cite{badminton-dataset-towards-olympic}, while maintaining comparable scale.

We further propose DSTA, which decouples spatio-temporal modeling into temporal, vertical, and horizontal branches to capture complementary motion cues for fine-grained action discrimination. Extensive experiments demonstrate that DSTA achieves state-of-the-art performance on both ShuttleSet and Fine-Badminton, with mAP improvements of 74.67\% and 66.23\%, respectively, while introducing only marginal computational overhead. Ablation studies validate the effectiveness of each component.

The proposed modeling strategy is general and applicable to other sports with similar fine-grained and rapid motion patterns, such as tennis and table tennis. It also enables a range of sports analytics applications, including highlight generation, tactical analysis, searchable video retrieval, and athlete training, and can be extended to real-time deployment with further optimization.

Overall, this work provides a new benchmark and an efficient spatio-temporal modeling framework for badminton video understanding. Future work will explore stronger backbones and integrate pose-based representations for more comprehensive analysis.

\bibliographystyle{IEEEtran}
\bibliography{ref}

\vspace{11pt}

\begin{IEEEbiographynophoto}{Tianyu Wang}
received the B.S. degree in Mathematics and Applied Mathematics and the M.S. degree in Electronic and Communication Engineering from Beihang University, Beijing, China, and the M.S. and Ph.D. degrees in Automation and Robotics and Information Technology from École Centrale de Nantes, Nantes, France. He was a Postdoctoral Researcher with the University of Toronto, Toronto, ON, Canada. He is currently an Associate Professor with the School of Economics and Management, Beihang University.
His research interests include optimization, scheduling theory, and data-driven decision making for intelligent systems.
\end{IEEEbiographynophoto}

\begin{IEEEbiographynophoto}{Junjie Wu}
received the Ph.D degree in management science and engineering from Tsinghua University. He is currently a full professor with Information Systems Department, Beihang University, and the director of Research Center for Data Intelligence. His research interests include data mining and complex networks. He was the recipient of NSFC Distinguished Young Scholars Award and the MOE Changjiang Young Scholars Award in China.
\end{IEEEbiographynophoto}

\begin{IEEEbiographynophoto}{Jingquan Gao}
received the B.B.A. degree in Industrial Engineering from Beihang University, Beijing, China, in 2025. He is currently pursuing the M.S. degree in Management Science and Engineering with the School of Economics and Management, Beihang University. His research interests include machine learning, multimedia data analysis, and optimization.
\end{IEEEbiographynophoto}

\begin{IEEEbiographynophoto}{Shishuo Li}
received the B.B.A. degree in Management from Beihang University, Beijing, China, in 2023. He is currently pursuing the M.S. degree in Management Science and Engineering with the School of Economics and Management, Beihang University. His research interests include machine learning, multimedia data analysis, and optimization.
\end{IEEEbiographynophoto}

\vfill

\end{document}